\documentclass{article}
\usepackage{spconf,amsmath,amssymb,amsfonts}
\usepackage{array}
\usepackage{graphicx}
\usepackage{booktabs}
\usepackage{multirow}
\usepackage{adjustbox}
\usepackage{pifont}
\usepackage{xcolor}
\usepackage{hyperref}

\makeatletter
\providecommand{\bmvaOneDot}{.}
\makeatother

\title{Lightweight RGB-T Tracking with Mobile Vision Transformers}

\name{Mahdi Falaki \qquad Maria A. Amer}
\address{Department of Electrical and Computer Engineering\\
Concordia University, Montr\'eal\\
Qu\'ebec, Canada\\
\texttt{mahdi.falaki@mail.concordia.ca, maria.amer@concordia.ca}}

\begin{document}
\maketitle

% --- your macros (verbatim definitions)
\def\eg{\emph{e.g}\bmvaOneDot}
\def\Eg{\emph{E.g}\bmvaOneDot}
\def\etal{\emph{et al}\bmvaOneDot}

% --------------------
% BODY (verbatim from bmvc_final.tex)
% --------------------
\begin{abstract}
Single-modality tracking (RGB-only) struggles under low illumination, weather, and occlusion. Multimodal tracking addresses this by combining complementary cues. While Vision Transformer-based trackers achieve strong accuracy, they are often too large for real-time. We propose a lightweight RGB-T tracker built on MobileViT with a progressive fusion framework that models intra- and inter-modal interactions using separable mixed attention. This design delivers compact, effective features for accurate localization, with under 4M parameters and real-time performance of 25.7 FPS on the CPU and 122 FPS on the GPU, supporting embedded and mobile platforms. To the best of our knowledge, this is the first MobileViT-based multimodal tracker. Model code and weights are available in \href{https://github.com/MahdiFalaki/Lightweight-RGB-T-object-tracker-using-Mobile-Vision-Transfomers}{the GitHub repository}.
\end{abstract}


\section{Introduction}
\label{sec:intro}
Object tracking aims to localize a target across video from an initial box. Trackers operate either on RGB alone or fuse RGB with additional modalities (thermal infrared, depth, events, language). RGB-only methods degrade under low illumination, adverse weather, and occlusion, motivating multimodal tracking that leverages complementary cues to overcome these challenges.

State-of-the-art multimodal trackers commonly adopt Vision Transformers (ViT)~\cite{vit}, achieving strong accuracy but incurring high computational cost that hinders real-time use. For example, SUTrack~\cite{sutrack}, which adopts a hierarchical transformer~\cite{hivit} backbone, remains heavyweight (L384 has $\sim$247M params, $\sim$12 FPS). We address efficiency by proposing a lightweight RGB–T tracker built on MobileViTv2~\cite{mobilevitv2}. The backbone combines local CNN processing with global attention, and we employ a progressive fusion scheme: separable mixed-attention performs template–search reasoning within each modality, followed by cross-modal fusion at deeper semantic layers. We adapt SMAT’s separable mixed attention\cite{smat}, which is derived from MobileViTv2’s separable self-attention~\cite{mobilevitv2}, and apply it from RGB-only to RGB–T, enabling compact intra- and inter-modal interactions.

Our contributions are summarized as follows:
\begin{itemize}
\item First RGB–T tracker built on MobileViTv2 for multimodal tracking.
\item Progressive fusion that couples intra-modal reasoning with late inter-modal interaction without extra backbone fusion blocks.
\item Model has $<4$M parameters and achieves inference speeds of 122 FPS on the GPU and 25.7 FPS on the CPU, with competitive accuracy compared to efficient multimodal state-of-the-art trackers.
\end{itemize}

\begin{figure*}[t]
\begin{center}
\hspace{-0.3cm}\includegraphics[width=\linewidth, height=4.5cm]{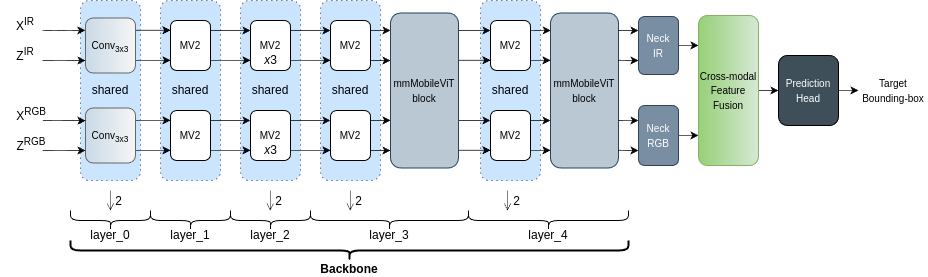}
\end{center}
\caption{The pipeline of proposed RGB-T tracker. MV2 stands for MobileNetV2 (Inverted Residual blocks) and mmMobileViT for multimodal MobileViT (see Figure~\ref{fig:mmMobileViT}). $\downarrow 2$ indicates spatial downsampling by 2. \{$X^{\mathrm{IR}}$,  $Z^{\mathrm{IR}}$\} show the input search and template frames of Thermal Infrared Modality (IR). \textit{$\times 3$} shows the number of subsequent MV2 blocks in layer\_2.}
\label{fig:pipeline}
\end{figure*}

\section{Related Work}

Multimodal trackers fall into two groups: RGB–T trackers specialized for thermal fusion \cite{ainet, stmt, tbsi, caformer} and RGB–X trackers targeting multiple modalities (RGB–D/E/T/L; depth, event, thermal infrared, and natural language)~\cite{sutrack, vipt, aptrack, emtrack}.

Architecturally, recent multimodal trackers are either single-stream transformer models \cite{vipt, tbsi, emtrack, ainet, sutrack} or Siamese networks \cite{siamfea, siamtdr}. ViT-based designs often fuse features within shared attention layers, whereas Siamese designs use separate modality streams and explicit/late fusion.

ViT \cite{vit} and HiViT \cite{hivit} backbones have advanced multimodal tracking but remain costly in terms of computation. The two most prominent trackers in this group are SUTrack~\cite{sutrack}, which offers unified RGB–X framework with a HiViT backbone.

In parallel, RGB-only tracking explored lightweight backbones to improve the accuracy–efficiency trade-off \cite{smat, hcat}, utilizing compact transformers such as MobileViTv2~\cite{mobilevitv2}. These advances are less adopted in multimodal tracking, where heavy backbones and complex fusion are common.

Recently, some multimodal trackers targeted efficiency, but they still face limitations. CMD \cite{cmd} (Knowledge Distillation model with ResNet18~\cite{resnet}) reduces cost but retains moderate size/speed; SiamTDR \cite{siamtdr} (AlexNet) lowers complexity at the expense of accuracy; TBSI \cite{tbsi} uses a small ViT with fused-template bridging, but introduces nontrivial training complexity.

\noindent An efficient RGB-X tracker, EMTrack \cite{emtrack}, adopts small ViT variants with modality experts, lightweight fusion, and multistage Knowledge Distillation, achieving good speed, but with relatively high parameter counts and complex training. SUTrack \cite{sutrack} also offers a compact HiViT variant, which is the state-of-the-art efficient multimodal tracker.

In general, previous work improves robustness but does not simultaneously achieve competitive accuracy, low parameter count, and high inference speed on resource-constrained devices, which motivates our design.

\section{Proposed Tracker}
In  this section, we overview our lightweight RGB–T tracker (Fig.~\ref{fig:pipeline}), composed of a mmMobileViT backbone (\ref{subsection:backbone}), a PW-XCorr neck (\ref{subsection:neck}), a cross-modal fusion transformer (\ref{subsec:crossfusion}), and a prediction head (\ref{subsec:head}).

%We present our training strategy with associated losses in Section \textcolor{red}{3.5}.

\subsection{Proposed Multimodal Mobile Vision Transformer-based Backbone}
\label{subsection:backbone}

Given template $Z$ and search $X$ from RGB and IR, we denote
$X_{\mathrm{in}}^{\mathrm{RGB/IR}}\!\in\!\mathbb{R}^{W_x\times H_x\times 3}$ and
$Z_{\mathrm{in}}^{\mathrm{RGB/IR}}\!\in\!\mathbb{R}^{W_z\times H_z\times 3}$.
A depthwise $3{\times}3$ convolution followed by a pointwise $1{\times}1$ projection increases the channel dimension while reducing spatial resolution. Modality-shared Inverted Residual blocks to produce spatially downsampled features, reducing redundancy while extracting modality-invariant cues for subsequent global relation modeling.

\begin{figure}
\begin{center}
\includegraphics[width=1.1\linewidth, height=5cm]{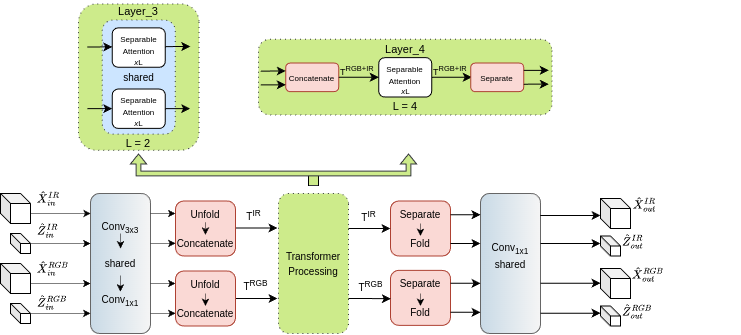}
\end{center}
\caption{mmMobileViT: Layer~3 uses intra-modal separable mixed attention; Layer~4 uses inter-modal. $L$: transformer layers per block.}
\label{fig:mmMobileViT}
\end{figure}

In Layer~3 (Fig.~\ref{fig:mmMobileViT}), the mmMobileViT block extends MobileViTv2~\cite{mobilevitv2} to multimodal tracking. After a shared $3{\times}3$ depthwise convolution and $1{\times}1$ projection ($C\!\rightarrow\!C_1$), features $\{\hat{X},\hat{Z}\}_{\text{in}}^{\{\mathrm{RGB},\mathrm{IR}\}}$ are partitioned into non-overlapping $p_1{\times}p_1$ patches and flattened to tokens of dimension $d$; with spatial size $H\!\times\!W$, the token count per map is $N=HW/p_1^2$. For each modality, template and search tokens are concatenated along the sequence dimension,
\begin{equation}
\begin{aligned}
\mathbf{T}^{\mathrm{RGB}} &= \big[\mathbf{Z}_{\mathrm{patch}}^{\mathrm{RGB}}
\parallel \mathbf{X}_{\mathrm{patch}}^{\mathrm{RGB}}\big]
\in \mathbb{R}^{C_1 \times N \times d},\\
\mathbf{T}^{\mathrm{IR}}  &= \big[\mathbf{Z}_{\mathrm{patch}}^{\mathrm{IR}}
\parallel \mathbf{X}_{\mathrm{patch}}^{\mathrm{IR}}\big]
\in \mathbb{R}^{C_1 \times N \times d}.
\end{aligned}
\end{equation}
A stack of $L$ shared transformer layers with separable mixed-attention~\cite{smat} models global dependencies of the concatenated template–search tokens within each modality. The factorized formulation operates with linear complexity $\mathcal{O}(Nd)$ in the token count $N$ and embedding dimension $d$, enabling strong context modeling at low cost. Tokens are folded back to feature maps and projected to channel size $C$. This layer performs intra-modal reasoning only; delaying cross-modal interaction preserves modality-specific structure and avoids premature mixing, as supported by ablations (Section \ref{subsec:ablation}).

Layer~4 follows the same tokenization pipeline (projection to $C_2$, patch size $p_1$) but concatenates RGB and IR sequences to enable cross-modal reasoning:
\begin{equation}
\mathbf{T}^{\mathrm{RGB+IR}}=\big[\mathbf{T}^{\mathrm{RGB}}\parallel \mathbf{T}^{\mathrm{IR}}\big]\in\mathbb{R}^{C_2\times 2N\times d}.
\end{equation}
Another $L$ layers of separable mixed-attention operate over $\mathbf{T}^{\mathrm{RGB+IR}}$ to capture inter-modal relations; outputs are folded to spatial maps and projected to produce fused features.

\begin{table*}[t]
\small
\centering
\renewcommand{\arraystretch}{1} % row height (slight reduction)
\setlength{\tabcolsep}{3pt}      % reduce horizontal padding
\begin{tabular}{l|c|c|c|ccc|cc|cc}
\toprule
\textbf{Tracker} & \textbf{\#Params} & \textbf{MACs} & \textbf{FPS} & 
\multicolumn{3}{c|}{\textbf{LasHeR}~\cite{lasher}} & 
\multicolumn{2}{c|}{\textbf{RGBT234}~\cite{rgbt234}} & 
\multicolumn{2}{c}{\textbf{GTOT}~\cite{gtot}} \\
& (M) & (G) & (GPU) & PR & SR & NPR & MPR & MSR & PR & SR \\
\midrule
SUTrack\_Tiny~\cite{sutrack} & 22   & 3   & \textcolor{green}{100}   & \textcolor{red}{0.667} & \textcolor{red}{0.539} & --   & \textcolor{red}{0.859} & \textcolor{red}{0.638} & 0.853 & \textcolor{blue}{0.726} \\
EMTrack~\cite{emtrack}       & 16   & 2   & \textcolor{blue}{83.8}  & \textcolor{green}{0.659} & \textcolor{green}{0.533} & --   & \textcolor{green}{0.838} & \textcolor{green}{0.601} & --    & --    \\
CMD~\cite{cmd}               & 19.9 & --  & 30    & 0.590 & 0.464 & \textcolor{blue}{0.546} & \textcolor{blue}{0.824} & 0.584 & \textcolor{green}{0.892} & \textcolor{green}{0.734} \\
TBSI\_Tiny~\cite{tbsi}       & 14.9 & --  & 40    & \textcolor{blue}{0.617} & \textcolor{blue}{0.489} & \textcolor{red}{0.578} & 0.794 & 0.555 & \textcolor{blue}{0.881} & 0.706 \\
\textbf{Ours}                & \textcolor{red}{3.93} & 4.35 & \textcolor{red}{121.9} & 0.603 & 0.473 & \textcolor{green}{0.567} & 0.806 & \textcolor{blue}{0.589} & \textcolor{red}{0.895} & \textcolor{red}{0.7467} \\
\midrule
SMAT*~\cite{smat}            & 3.76 & -- & 154.6 & 0.549 & 0.438 & 0.512 & 0.737 & 0.536 & 0.690 & 0.578 \\
\bottomrule
\end{tabular}
\caption{Comparison on LasHeR~\cite{lasher}, RGBT234~\cite{rgbt234}, and GTOT~\cite{gtot}. Best/second/third are \textcolor{red}{red}/\textcolor{green}{green}/\textcolor{blue}{blue}. \textbf{SMAT*}~\cite{smat}: RGB-only baseline trained on LasHeR. Parameters and MAC counts are reported in millions (M) and gigas (G), respectively.}
\label{tab:table1}
\end{table*}

\vspace{-0.75\baselineskip}

\subsection{Neck Module}
\label{subsection:neck}
We fuse template and search per modality using pixel-wise cross-correlation (PW-XCorr) as in \cite{smat}, yielding $\mathbf{F}^{m}=\mathrm{PWCorr}(\mathbf{X}^{m},\mathbf{Z}^{m})$ with $C_f$ channels ($m\!\in\!\{\mathrm{RGB},\mathrm{IR}\}$) for the fusion transformer.

\subsection{Proposed Cross-Modal Fusion Transformer}
\label{subsec:crossfusion}
RGB and IR neck features are patchified (size $p_2$), flattened, concatenated, and processed by $L$ layers of separable mixed attention to model global inter-modal relations. The attended maps are merged via channel-wise weighted addition:
\begin{equation}
\mathbf{F}_{\mathrm{fused}}=\sigma(\mathbf{W}^{\mathrm{RGB}})\odot \mathbf{F}^{\mathrm{RGB}}
+\sigma(\mathbf{W}^{\mathrm{IR}})\odot \mathbf{F}^{\mathrm{IR}},
\end{equation}
with learnable weights of $\mathbf{W}^{\mathrm{RGB}},\mathbf{W}^{\mathrm{IR}}\!\in\!\mathbb{R}^{1\times C_f\times1\times1}$.

\subsection{Prediction Head}
\label{subsec:head}
We adopt the SMAT~\cite{smat} head with parallel classification and regression branches (composed of $3{\times}3$ convolution, separable self-attention, final $3{\times}3$ convolution).

\subsection{Training Loss Function}
We use focal loss $L_{\mathrm{cls}}$ and $\ell_1$+GIoU regression losses $L_1,L_{\mathrm{giou}}$; the total loss is:
\begin{equation}
L_{\mathrm{total}} = L_{\mathrm{cls}} + \lambda_1 \cdot L_1 + \lambda_2 \cdot L_{\mathrm{giou}},
\label{eq:5}
\end{equation}
where $\lambda_1$, $\lambda_2$ balance contributions of the regression losses.

\section{Implementation Details and Experimental Results}

\subsection{Implementation Details}
\textbf{Model.} Input template/search sizes are $128{\times}128$/$256{\times}256$. Feature maps are progressively downsampled through four $2\times$ operations, yielding output spatial sizes of $8\times8$ and $16\times16$ for the template and search branches, respectively. mmMobileViT uses $L{=}2$ stacks in L3 and $L{=}4$ in L4; backbone patch size $p_1{=}2$, fusion transformer patch size $p_2{=}8$; the fusion transformer uses $C_f{=}128$ and one block. Across the backbone, the number of channels progresses as $\{3 \rightarrow 32 \rightarrow 64 \rightarrow 128 \rightarrow 256 \rightarrow 384\}$.

\textbf{Training.} Trained on LasHeR~\cite{lasher} for 60 epochs with AdamW (lr $4\times10^{-4}$, backbone lr$\times$0.1), step decay at epoch 40, batch 128. We initialize from MobileViTv2~\cite{mobilevitv2}, omit positional embeddings as in \cite{smat}, and use horizontal flip and brightness jitter for augmentations.

\textbf{Inference.} All inference results are generated on a machine equipped with an Intel Core i9-12900KF CPU and a single NVIDIA RTX 3090 GPU.

\begin{figure*}
\begin{center}
\hspace{-0.3cm}\includegraphics[width=\linewidth, height=6cm]{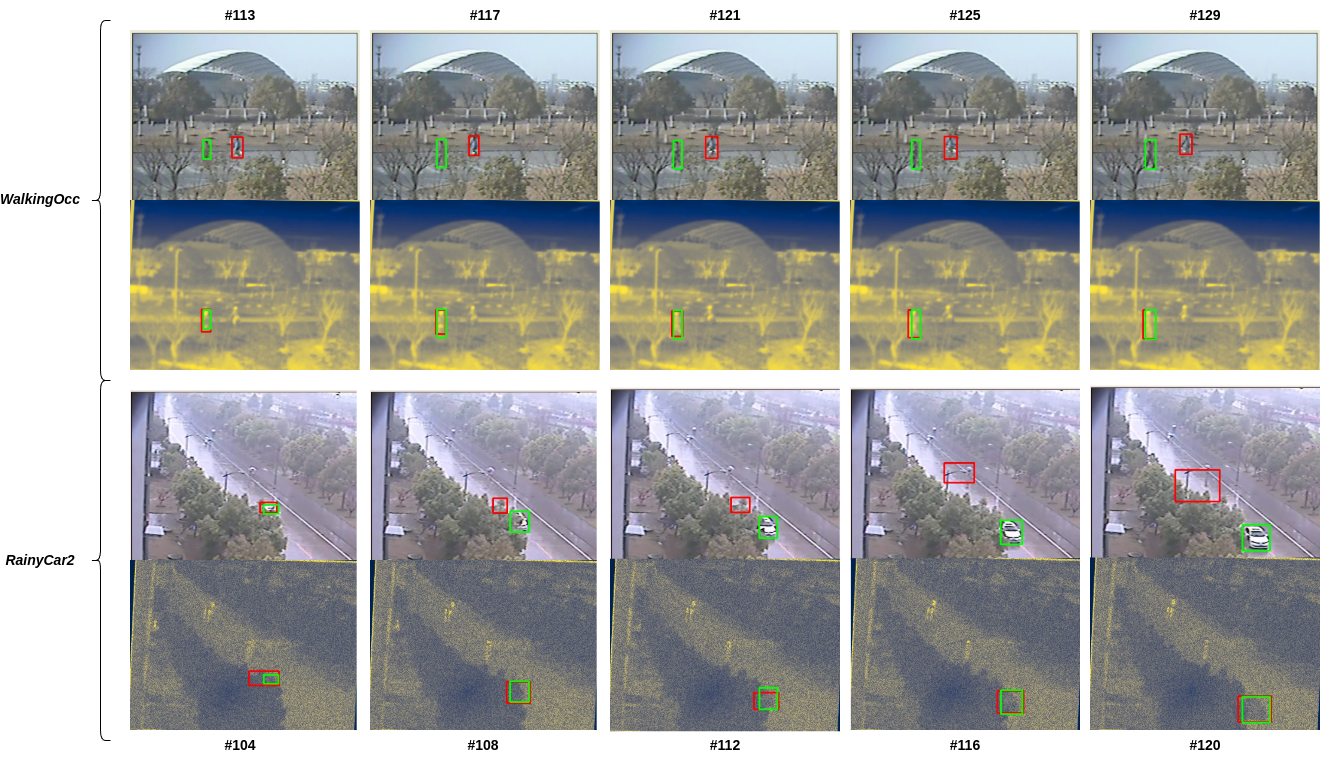}
\end{center}
\caption{Tracking on two GTOT~\cite{gtot} sequences comparing RGB-only (upper) and RGB-T (lower), as in Table 3. \textcolor{red}{Red}: predictions; \textcolor{green}{green}: ground truth. RainyCar2: rainy weather; WalkingOcc: partial occlusion.}
\label{fig:ablation}
\end{figure*}

\subsection{Comparison to Related Work}

We follow the standard RGB-T protocol: train on LasHeR~\cite{lasher} and test on LasHeR test, RGBT234~\cite{rgbt234}, and GTOT~\cite{gtot} using their official metrics (SR/PR/NPR, MSR/MPR, SR/PR).

We evaluate our tracker against recent efficient multimodal trackers, including SUTrack-Tiny~\cite{sutrack}, EMTrack~\cite{emtrack}, CMD~\cite{cmd}, and TBSI-Tiny~\cite{tbsi}. As shown in Table~\ref{tab:table1}, our model attains the fastest inference speed (FPS) and the fewest parameters, while maintaining competitive tracking accuracy.

On LasHeR and RGBT234, SUTrack-Tiny achieves the highest accuracy overall, while our tracker offers competitive accuracy with over 5.6× fewer parameters (22M vs.\ 3.93M) and about 20 higher FPS (100 vs.\ 121.9).

On GTOT, our tracker ranks best in PR and SR, indicating stronger overlap consistency for small targets~\cite{gtot}, while remaining more efficient than other efficient trackers. Relative to the RGB-only SMAT* baseline~\cite{smat}, complementary RGB–IR fusion via separable mixed-attention yields consistent accuracy gains with a modest speed trade-off.

Overall, our method consistently delivers a better trade-off across all datasets. With the fewest parameters (3.93M), the highest inference speed (121.9 FPS), and competitive tracking accuracy, our tracker is well-suited for real-time RGB-T applications on resource-constrained devices.

% =========================
% ABLATION STUDIES (FINAL)
% =========================
\subsection{Ablation Studies}
\label{subsec:ablation}

\textbf{Effect of Progressive Intra- and Inter-Modal Fusion.} We compare backbone fusion variants. \textit{No-Fusion}: both Layer~3 and Layer~4 are intra-modal (template--search reasoning within modality only). \textit{All-Fusion}: makes both layers inter-modal. \textit{Proposed} follows our progressive design.

\begin{table}
\centering
\footnotesize
\setlength{\tabcolsep}{2pt}
\renewcommand{\arraystretch}{1.12}
\begin{tabular*}{\linewidth}{@{\extracolsep{\fill}} l|c|ccc|cc|cc}
\toprule
\textbf{Variant} & \textbf{FPS} & \multicolumn{3}{c|}{\textbf{LasHeR}} & \multicolumn{2}{c|}{\textbf{RGBT234}} & \multicolumn{2}{c}{\textbf{GTOT}} \\
 & (GPU) & PR & nPR & SR & MPR & MSR & PR & SR \\
\midrule
Proposed   & 121.92 & \textbf{0.603} & \textbf{0.567} & \textbf{0.473} & \textbf{0.8063} & \textbf{0.5890} & \textbf{0.8949} & \textbf{0.7467} \\
No-Fusion  & 110.00 & 0.589 & 0.556 & 0.467 & 0.7824 & 0.5652 & 0.8612 & 0.7171 \\
All-Fusion & 129.00 & 0.584 & 0.543 & 0.450 & 0.7812 & 0.5714 & 0.8639 & 0.7077 \\
\bottomrule
\end{tabular*}
\caption{Fusion placement inside the backbone (Fig.~\ref{fig:mmMobileViT}).}
\label{tab:ablation_progressive_fusion}
\end{table}

Table~\ref{tab:ablation_progressive_fusion} shows consistent gains for \textit{Proposed} across all datasets. The \textit{No-Fusion} variant confirms that some explicit cross-modal interaction is necessary. On the other hand, the \textit{All-Fusion} performance indicates that early mixing harms discriminability by disrupting modality-specific structure before sufficient intra-modal reasoning. This directly supports the Section~\ref{subsection:backbone} assumption: deferring RGB-IR fusion until deeper features yields stronger representations.

% ---- (2) Modal contribution + fusion transformer
\textbf{Modal Contribution and Fusion Transformer.} To isolate the contributions of the IR stream and the cross-fusion transformer (Section~\ref{subsec:crossfusion}), we evaluate two reduced variants on RGBT234~\cite{rgbt234} and GTOT~\cite{gtot} (Table~\ref{tab:ablation}). SMAT* is the RGB-only baseline (same as in Table~\ref{tab:table1}), with no IR stream and no fusion transformer. Variant~2 is dual-modality but removes the transformer layers from the fusion module, keeping only the weighted addition.

Figure~\ref{fig:ablation} provides qualitative visual examples from GTOT, showing how the RGB-T model overcomes the challenging conditions for RGB-only tracking.

\begin{table}
\centering
\footnotesize
\setlength{\tabcolsep}{3pt}
\renewcommand{\arraystretch}{1}
\begin{tabular*}{\linewidth}{@{\extracolsep{\fill}}
 l
 @{\hspace{5pt}}|@{\hspace{0.1pt}}
 c
 @{\hspace{5pt}}|@{\hspace{0.1pt}}
 c
 @{\hspace{5pt}}|@{\hspace{0.1pt}}
 c c
 @{\hspace{5pt}}|@{\hspace{0.1pt}}
 c c
}
\toprule
\textbf{Model Variant} & \textbf{\#Params} & \textbf{FPS} &
\multicolumn{2}{c|}{\textbf{RGBT234}~\cite{rgbt234}} &
\multicolumn{2}{c}{\textbf{GTOT}~\cite{gtot}} \\
 & (M) & (GPU) & MPR & MSR & MPR & MSR \\
\midrule
SMAT*         & 3.767 & 154.64 & 0.7378 & 0.5364 & 0.6904 & 0.5785 \\
Variant 2     & 3.786 & 124.00 & 0.7860 & 0.5704 & 0.8318 & 0.6902 \\
Proposed & 3.926 & 121.92 & \textbf{0.8063} & \textbf{0.589} & \textbf{0.8838} & \textbf{0.7409} \\
\bottomrule
\end{tabular*}
\caption{Comparing SMAT* (RGB-only), Variant~2 (dual-modal w/o cross-fusion transformer), and the full model. SMAT* is the same baseline as in Table~\ref{tab:table1}.}
\label{tab:ablation}
\end{table}

\section{Conclusion}

We presented a lightweight RGB-T tracker built on MobileViTv2 with progressive intra-to-inter modality fusion via separable mixed attention and a final cross-modal fusion transformer. The design attains a strong accuracy--efficiency trade-off ($<4$M parameters, 122 FPS on GPU, 25.7 FPS on CPU) while maintaining competitive tracking quality. Ablations show that incorporating the thermal stream and the architectural choices in fusion are both necessary for gains.

A current limitation is the token growth from modality concatenation, which modestly reduces speed; efficient token pruning could mitigate this. Moreover, extending the framework to other modalities, such as RGB-D, to assess the model's performance is a valuable next experiment.

\clearpage
\bibliographystyle{IEEEbib}
\bibliography{egbib}
\end{document}